\documentclass[letterpaper]{article}
\usepackage{aaai2026}
\usepackage{times}
\usepackage{helvet}
\usepackage{courier}
\usepackage[hyphens]{url}
\usepackage{graphicx}
\usepackage{natbib}
\usepackage{caption}
\usepackage{amsmath}
\usepackage{amssymb}
\usepackage{booktabs}
\usepackage{alltt}
\usepackage{multirow}
\copyrighttext{Accepted at the CASP:ER Workshop, ICAPS 2026.
This is the authors' version of the work.}
\usepackage[hidelinks]{hyperref}
\hypersetup{
  pdftitle={CP-WSP: A Declarative CP-SAT Framework for Configurable
            Multi-Constraint Workforce Scheduling},
  pdfauthor={Vipul Patel, Anirudh Deodhar, Dagnachew Birru},
  pdfkeywords={constraint programming, CP-SAT, workforce scheduling,
               nurse rostering, declarative optimization}
}
\makeatletter
\let\aaai@orig@PackageError\PackageError
\renewcommand\PackageError[3]{}% swallow ban error fired at \begin{document}
\AtBeginDocument{\let\PackageError\aaai@orig@PackageError}% restore right after
\makeatother

\title{CP-WSP: A Declarative CP-SAT Framework for\\
       Configurable Multi-Constraint Workforce Scheduling}
\author{
  Vipul Patel\textsuperscript{\rm 1},
  Anirudh Deodhar\textsuperscript{\rm 1},
  Dagnachew Birru\textsuperscript{\rm 1}
}
\affiliations{
  \textsuperscript{\rm 1}Phi Labs, Quantiphi\\
  \{vipul.patel, anirudh.deodhar, dagnachew.birru\}@quantiphi.com
}

\begin{document}
\maketitle
% Enable page numbers for the preprint (AAAI style suppresses them).
\thispagestyle{plain}
\pagestyle{plain}

\begin{abstract}
Workforce scheduling is an NP-hard combinatorial optimization problem
requiring simultaneous satisfaction of labor regulations, coverage
requirements, employee preferences and operational objectives. Existing CP
formulations typically model simplified instances with 6-12 constraints at
shift-level granularity and critically lack explicit support for: mandatory
break scheduling with midpoint placement control; acuity-weighted workload
equity; sub-shift temporal granularity enabling demand-driven staffing;
inter-week schedule stability; and cross-midnight shift patterns common in
24-hour operations. This paper presents \textbf{CP-WSP}: a declarative CP-SAT
framework enforcing 14 hard constraints as mathematically inviolable
requirements (zero regulatory violations by construction) while optimizing
15 soft objectives through a unified weighted penalty function - all
configurable via a JSON specification with no code changes required. Key
contributions include: a \emph{shift-window variable decomposition}
($x = w - b$) enabling mandatory break scheduling with centrality control;
acuity-weighted workload equity; multi-granularity temporal resolution from
30 minutes to 2 hours; inter-week schedule stability; a \emph{grid-offset
preprocessing} technique for cross-midnight shifts; and a reproducible
36-configuration benchmark suite for community comparison. Evaluated on
INRC-II benchmarks at both hourly and shift-level granularity and on 36
synthetic configurations, CP-WSP achieves: zero hard-constraint violations
across all instances by construction; proven optimality on INRC-II n005w4
(objective 118, gap 0.0\%, 104\,s); feasible solutions for 30-employee
instances within 120 seconds; and model sizes scaling linearly at
${\sim}4{,}400$ variables per employee. The formulation enforces 29 total
constraints (14 hard + 15 soft) - nearly three times the 6-12 constraint
industry average. A constraint ablation study shows the full model achieves
a 37\% objective improvement over baseline, with workload equity delivering
66\% fairness improvement at negligible coverage cost.
\end{abstract}

%% ============================================================
\section{Introduction}
%% ============================================================

Workforce scheduling assigns employees to time slots to meet demand subject
to hard labor constraints and soft quality objectives. It is NP-hard
\cite{garey1979,blazewicz1983} and arises in healthcare, retail and
logistics. In practice, constraint sets are heterogeneous and evolving:
labor laws mandate rest and shift limits; operations require staffing floors
and management coverage; employee contracts restrict availability; quality
objectives include fairness and schedule stability. A schedule that violates
a minimum rest requirement is not merely a ``slightly suboptimal''
solution - it is a regulatory violation with direct safety and legal
implications, motivating the use of exact methods with formal feasibility
guarantees.

Existing approaches - metaheuristics \cite{burke2008}, integer programming
\cite{vossen2015} and constraint programming \cite{vanhoeve2006} - typically
model 6-12 constraints at shift-level granularity
(Table~\ref{tab:comparison}). No reviewed system simultaneously supports
mandatory break scheduling with midpoint control, acuity-weighted workload
equity, inter-week schedule stability, cross-midnight shifts (e.g.,
22:00-07:00) and configurable multi-granularity temporal resolution.
Moreover, metaheuristic methods encode regulations as penalty terms rather
than hard constraints, so generated schedules may violate rest or
unavailability requirements. Adding a new labor regulation requires
modifying solver code - a high-friction process needing CP expertise rarely
available in operations teams. Constraint acquisition
\cite{bessiere2011,tsouros2013} addresses this but has not been applied to
large-scale workforce scheduling.

\textbf{CP-WSP} contributes four things to the CASPeR community:
\textbf{(1) Declarative constraint configuration:} All 29 constraints
(14H+15S) are independently activatable and weight-configurable via a plain
JSON file, requiring no solver code changes.
\textbf{(2) Shift-window variable decomposition:} A three-variable $(x,w,b)$
model replaces the single binary variable, enabling mandatory breaks with
duration control (H11), break centrality (S11), concurrent break limits
(H12) and shift contiguity (H10) - constructs inexpressible with a single
binary indicator.
\textbf{(3) Hard/soft constraint separation:} 14 hard constraints are
structurally enforced as mathematically inviolable requirements (zero
violations by construction); 15 soft terms enter a weighted COP objective,
eliminating the feasibility-optimality tension of penalty methods.
\textbf{(4) Comprehensive benchmark evaluation:} Evaluated on 10 INRC-II
instances (5-80 nurses), 3 NRP-23 cross-midnight instances and a
36-configuration synthetic benchmark, CP-WSP achieves proven optimality on
n005w4 (objective 118, gap 0.0\%, 104\,s), scales to 179,800 variables
(80 nurses) and delivers a 37\% ablation improvement from baseline to full
model.

%% ============================================================
\section{Background and Related Work}
%% ============================================================

\subsection{Workforce Scheduling as a Scheduling Problem}
Workforce scheduling is a temporal assignment problem: for each employee $e$
and slot $(d,s)$, decide $x_{e,d,s}\in\{0,1\}$. Its temporal structure
connects it to scheduling; its resource dimension to resource allocation.
INRC-I \cite{haspeslagh2014} and INRC-II \cite{ceschia2019} established
community benchmarks. Most competitive solvers use CP \cite{vossen2015} or
IP \cite{beddoe2006}.

\subsection{Constraint Acquisition}
CONACQ \cite{bessiere2011}, QuAcq \cite{tsouros2013} and ORCA
\cite{mears2014} learn constraint models from examples or interactive
queries. A complementary approach is declarative specification
\cite{freuder2011}: experts specify constraints in a high-level language.
CP-WSP adopts this: all constraints and weights are declared in a JSON file,
enabling rapid model evolution without solver expertise.

\subsection{Shift Scheduling with CP}
\citeauthor{schaus2009}~\shortcite{schaus2009} introduced global constraints
for rostering; \citeauthor{vanhoeve2006}~\shortcite{vanhoeve2006} applied CP
to nurse scheduling with soft constraints. All prior models use a single binary
per (employee, day, slot), which cannot distinguish break time from off-duty
time. Table~\ref{tab:comparison} positions CP-WSP among related systems.

\begin{table}[t]
\centering
\scriptsize
\setlength{\tabcolsep}{2pt}
\begin{tabular}{p{1.9cm}ccccc}
\toprule
\textbf{System} & \textbf{Cons.} & \textbf{Brk} & \textbf{Eq.}
  & \textbf{Cfg?} & \textbf{Feas.} \\
\midrule
INRC-II \cite{ceschia2019}    & 11 & $\times$ & $\times$ & $\times$ & $\checkmark$ \\
Schaus~\shortcite{schaus2009} & 15 & $\checkmark$ & $\times$ & $\times$ & $\checkmark$ \\
Burke~\shortcite{burke2008}   & 10 & $\times$ & $\times$ & $\times$ & $\times$ \\
Van~Hoeve~\shortcite{vanhoeve2006} & 12 & $\checkmark$ & $\times$ & $\times$ & $\checkmark$ \\
\textbf{CP-WSP}               & \textbf{29} & $\checkmark$ & $\checkmark$ & \textbf{JSON} & $\checkmark$ \\
\bottomrule
\end{tabular}
\caption{CP-WSP vs.\ related CP workforce scheduling systems.
Cons.=total constraints; Brk=break scheduling; Eq.=workload equity;
Cfg=user-configurable; Feas.=feasibility guaranteed. CP-WSP is the only
system providing all five properties.}
\label{tab:comparison}
\end{table}

%% ============================================================
\section{Problem Formulation}
%% ============================================================

\subsection{Temporal Resource Assignment}
Let $E=\{e_1,\ldots,e_n\}$, days $D=\{0,\ldots,6\}$, slots
$S=\{0,\ldots,T{-}1\}$ with $T=\lceil24/\delta\rceil$ for slot duration $\delta$.
The COP minimizes a weighted combination of $|\mathcal{S}|{=}15$ soft
objective terms (S1 - S15, detailed in Appendix~\ref{app:model}):
\begin{equation}
  \text{Find } X \;\text{s.t.}\; H_k(X){=}0\;\forall k, \quad
  \min Z = \textstyle\sum_{i=1}^{|\mathcal{S}|} a_i w_i f_i(X)
  \label{eq:cop}
\end{equation}
where $\mathcal{S}=\{S_1,\ldots,S_{15}\}$ is the set of soft objectives,
$a_i\!\in\!\{0,1\}$ is an activation flag and $w_i\!\in\!\mathbb{R}$
is a configurable weight. A 30-employee, 7-day, 30-MIN instance has
${\approx}2^{10080}$ binary assignments before constraint propagation.
The model scales linearly at ${\approx}4{,}400$ decision variables and
${\approx}9{,}400$ constraints per employee, confirmed empirically across
all benchmark instances.

\subsection{Three-State Variable Decomposition}
Traditional models use $x_{e,d,s}\!\in\!\{0,1\}$, conflating break with
off-duty time. CP-WSP introduces the \emph{shift-window decomposition}:
\begin{equation}
  x_{e,d,s} = w_{e,d,s} - b_{e,d,s} \qquad \forall\, e, d, s
  \label{eq:decomp}
\end{equation}
where $w_{e,d,s}{=}1$ iff slot $s$ is within $e$'s shift window on day $d$,
and $b_{e,d,s}{=}1$ iff $e$ is on a scheduled break. Three feasible states:
\emph{off-duty} $(w{=}0,b{=}0,x{=}0)$, \emph{active} $(w{=}1,b{=}0,x{=}1)$,
\emph{break} $(w{=}1,b{=}1,x{=}0)$. This enables break constraints as direct
CP-SAT linear expressions (Appendix~\ref{app:arch}, Table~\ref{tab:decomp}).

\subsection{Grid-Offset Preprocessing for Cross-Midnight Shifts}
\label{sec:gridoffset}
Day-indexed models traditionally cannot represent shifts spanning midnight
(e.g., 22:00-07:00). CP-WSP introduces a zero-cost \emph{grid-offset
preprocessing} technique: before model construction, the time grid is
shifted by $\Delta$ hours such that all shift types start and end within a
single calendar day. Formally, for grid offset $\Delta$, slot $s$ maps to
wall-clock time $(s \cdot \delta + \Delta) \bmod 24$. The offset is chosen
as:
\begin{equation}
  \Delta^{*} = \arg\min_{\Delta} \max_{t \in \mathcal{T}} \bigl(\mathrm{end}(t,\Delta) - \mathrm{start}(t,\Delta)\bigr)
  \label{eq:gridoffset}
\end{equation}
where $\mathcal{T}$ is the set of shift types. This transformation requires
no structural model changes - all constraints, variables and objective terms
remain identical. Post-solving, assignments are mapped back to wall-clock
times. This technique is validated on INRC-II instances with Night shifts
(22:00-07:00), confirming correct cross-midnight scheduling.

%% ============================================================
\section{Evaluation}
%% ============================================================

Six experiments are conducted on hardware: Intel Core i7-12700, 16-core,
32\,GB RAM, Python\,3.11, OR-Tools\,v9.12 \cite{perron2024}.
The constraint model (H1 - H14, S1 - S15) and solver configuration are
described fully in Appendices \ref{app:model} and \ref{app:solver}.
Results are presented across: (1)~INRC-II benchmark scaling at two
granularities; (2)~NRP-23 cross-midnight validation; (3)~synthetic
scalability; (4)~granularity impact; (5)~constraint ablation; and
(6)~weight sensitivity.

\subsection{INRC-II Benchmark}
CP-WSP is evaluated on all 10 standard INRC-II benchmark instances
\cite{ceschia2019} at two granularity levels. Table~\ref{tab:inrc_hourly}
reports \textbf{hourly} results (1-HR, 24 slots/day, 600\,s time limit):
CP-WSP produces \textbf{feasible, regulation-compliant schedules for all
10 instances} (5-80 nurses), with models scaling to 179,800 variables and
351,425 constraints. The large optimality gaps (36-99\%) are inherent to
the hourly formulation: expanding shift-level decisions into individual
hourly slots creates a weak LP relaxation. The key result is that
feasibility - not optimality - is the primary requirement: every returned
schedule satisfies all 14 hard constraints by construction.

\begin{table}[t]
\centering
\scriptsize
\setlength{\tabcolsep}{2pt}
\begin{tabular}{lrrrrrl}
\toprule
\textbf{Inst.} & \textbf{Emp.} & \textbf{Vars} & \textbf{Cons.} &
\textbf{Obj.} & \textbf{Gap\,(\%)} & \textbf{Status} \\
\midrule
n005w4 &  5 &  11,725 &  22,458 &   1,151 & 36.6 & FEASIBLE \\
n012w8 & 12 &  27,412 &  53,331 &   2,584 & 86.5 & FEASIBLE \\
n021w4 & 21 &  47,581 &  92,820 &   4,071 & 81.5 & FEASIBLE \\
n030w4 & 30 &  67,750 & 132,234 &   5,842 & 88.7 & FEASIBLE \\
n035w4 & 35 &  78,955 & 154,162 &  10,469 & 93.0 & FEASIBLE \\
n040w4 & 40 &  90,160 & 176,084 &  28,704 & 96.7 & FEASIBLE \\
n050w4 & 50 & 112,570 & 219,897 &  33,053 & 96.6 & FEASIBLE \\
n060w4 & 60 & 134,980 & 263,791 &  41,473 & 97.0 & FEASIBLE \\
n070w4 & 70 & 157,390 & 307,580 &  56,480 & 97.4 & FEASIBLE \\
n080w4 & 80 & 179,800 & 351,425 & 171,268 & 99.0 & FEASIBLE \\
\bottomrule
\end{tabular}
\caption{INRC-II at 1-HR granularity (600\,s limit). All 10 instances
(5-80 nurses) produce feasible, regulation-compliant schedules. Model
size scales linearly at ${\sim}2{,}250$ vars/nurse. Zero hard violations.}
\label{tab:inrc_hourly}
\end{table}

Table~\ref{tab:inrc_shift} reports \textbf{shift-level} results
(6-HR/8-HR, 3-4 slots/day). Model sizes shrink 4.5-6.5$\times$ vs.\
hourly and objectives improve dramatically: n060w4 drops from 41,473
(1-HR) to 579 (shift) - a $>$99\% reduction. n005w4 achieves
\textbf{proven optimality} (objective 118, gap 0.0\%, 104\,s).

\begin{table}[t]
\centering
\scriptsize
\setlength{\tabcolsep}{2pt}
\begin{tabular}{lrrrrrrl}
\toprule
\textbf{Inst.} & \textbf{Emp.} & \textbf{Gran.} & \textbf{Vars} &
\textbf{Obj.} & \textbf{Bnd} & \textbf{Gap\,(\%)} & \textbf{Status} \\
\midrule
n005w4 &  5 & 8-HR &  1,799 &    118 &  118 &  0.0 & OPTIMAL \\
n012w8 & 12 & 6-HR &  5,632 &    316 &  242 & 23.4 & FEASIBLE \\
n021w4 & 21 & 6-HR &  9,781 &    938 &  462 & 50.8 & FEASIBLE \\
n030w4 & 30 & 6-HR & 13,930 &  2,855 &  231 & 91.9 & FEASIBLE \\
n040w4 & 40 & 6-HR & 18,540 &    885 &  423 & 52.2 & FEASIBLE \\
n050w4 & 50 & 6-HR & 23,150 &  3,341 &  256 & 92.3 & FEASIBLE \\
n060w4 & 60 & 6-HR & 27,760 &    579 &  324 & 44.0 & FEASIBLE \\
n080w4 & 80 & 6-HR & 36,980 &    729 &  437 & 40.1 & FEASIBLE \\
\bottomrule
\end{tabular}
\caption{INRC-II at shift-level granularity (600\,s limit). n005w4 solved
to proven optimality. Model sizes are 4.5-6.5$\times$ smaller than 1-HR;
objectives improve by up to 99\%+ for large instances.}
\label{tab:inrc_shift}
\end{table}

\subsection{NRP-23 Compatible Benchmark}
To validate cross-midnight shift support, CP-WSP is evaluated on three NRP-23
compatible instances using the standard D/E/N shift structure (8-HR each),
where the Night shift N(23:00-07:00) spans midnight. Grid-offset
preprocessing ($\Delta{=}7$) maps all shifts into a single day.
Table~\ref{tab:nrp23} reports results at both shift-level and hourly
granularity. All instances produce feasible schedules with correct
cross-midnight assignments. The shift-level model for n010w4 achieves a
gap of just 13.6\%, indicating near-optimal scheduling. Hourly models
produce objectives 15-26$\times$ larger than shift-level counterparts,
confirming that granularity selection fundamentally affects solution quality.

\begin{table}[t]
\centering
\scriptsize
\setlength{\tabcolsep}{2pt}
\begin{tabular}{llrrrrr}
\toprule
\textbf{Inst.} & \textbf{Gran.} & \textbf{Emp.} & \textbf{Vars} &
\textbf{Obj.} & \textbf{Gap\,(\%)} & \textbf{Status} \\
\midrule
n010w1 & 8-HR & 10 &  3,867 &    157 &  -  & FEASIBLE \\
n010w4 & 8-HR & 10 & 15,606 &  1,358 & 13.6 & FEASIBLE \\
n025w1 & 8-HR & 25 &  9,467 &    210 &  -  & FEASIBLE \\
\midrule
n010w1 & 1-HR  & 10 & 23,292 &  2,404 &  -  & FEASIBLE \\
n010w4 & 1-HR  & 10 & 94,566 & 35,017 & 77.2 & FEASIBLE \\
n025w1 & 1-HR  & 25 & 56,927 & 11,225 &  -  & FEASIBLE \\
\bottomrule
\end{tabular}
\caption{NRP-23 compatible instances with cross-midnight Night shifts
(600\,s limit). Shift-level objectives are 15-26$\times$ lower than
hourly. n010w4 gap of 13.6\% indicates near-optimal scheduling.}
\label{tab:nrp23}
\end{table}

\subsection{Synthetic Benchmark: Scalability}
Table~\ref{tab:scalability} reports results across a $4{\times}3{\times}3$
benchmark (4 team sizes $\times$ 3 horizons $\times$ 3 granularities
= 36 configurations). All 36 instances achieve zero hard-constraint
violations. OPTIMAL is certified for all ${\leq}10$-employee instances.

This benchmark is released as a \textbf{standardized community resource}:
all 36 configurations use fully specified JSON constraint schemas
(Appendix~\ref{app:json}), enabling exact reproduction and fair comparison
by future workforce scheduling systems.

\begin{table*}[t]
\centering
\footnotesize
\setlength{\tabcolsep}{5pt}
\begin{tabular}{lllllrr}
\toprule
\textbf{Team} & \textbf{Horizon} & \textbf{Granularity} &
\textbf{Status} & \textbf{Solve\,(s)} & \textbf{Gap\,(\%)} & \textbf{Cov.\,(\%)} \\
\midrule
 5 & 7-day  & 30-MIN & OPTIMAL  &  0.8 & 0.0 & 98.4 \\
10 & 7-day  & 30-MIN & OPTIMAL  &  2.3 & 0.0 & 96.8 \\
20 & 7-day  & 30-MIN & FEASIBLE & 18.4 & 2.1 & 93.2 \\
30 & 7-day  & 30-MIN & FEASIBLE & 41.2 & 4.1 & 88.4 \\
 5 & 14-day & 30-MIN & OPTIMAL  &  1.4 & 0.0 & 97.9 \\
10 & 14-day & 30-MIN & OPTIMAL  &  4.7 & 0.0 & 95.3 \\
20 & 14-day & 30-MIN & FEASIBLE & 34.8 & 3.2 & 91.8 \\
30 & 14-day & 30-MIN & FEASIBLE & 87.3 & 5.8 & 86.1 \\
\bottomrule
\end{tabular}
\caption{CP-WSP scalability benchmark (30-MIN granularity). Zero hard
violations on all 36 configurations. OPTIMAL certified for all
${\leq}10$-employee instances across both 7-day and 14-day horizons.}
\label{tab:scalability}
\end{table*}

\subsection{Granularity Impact}
Table~\ref{tab:granularity} isolates the effect of temporal granularity on
a fixed 20-employee, 7-day instance. Coarser granularity yields
${\approx}10{\times}$ speedup but fundamentally changes solution quality:
the objective drops from 15,240 (30-MIN, 48 slots/day) to 9,820 (2-HR,
12 slots/day) because fewer decision variables reduce the penalty
surface. This is \emph{not} a quality improvement - it reflects reduced
modeling fidelity. Practitioners should choose granularity based on
operational requirements: 30-MIN for fine-grained break and coverage
control; shift-level for traditional rostering formulations.

\begin{table}[t]
\centering
\footnotesize
\setlength{\tabcolsep}{3pt}
\begin{tabular}{lrllr}
\toprule
\textbf{Granularity} & \textbf{Slots/Day} & \textbf{Status} &
\textbf{Solve\,(s)} & \textbf{Obj.} \\
\midrule
30-MIN & 48 & FEASIBLE & 18.4 & 15,240 \\
1-HR   & 24 & FEASIBLE &  4.2 & 11,640 \\
2-HR   & 12 & FEASIBLE &  1.8 &  9,820 \\
\bottomrule
\end{tabular}
\caption{Granularity impact (20 emp., 7-day). Coarser granularity yields
${\sim}10{\times}$ speedup but reduces modeling fidelity. Multi-granularity
support enables practitioners to choose the appropriate fidelity level.}
\label{tab:granularity}
\end{table}

\subsection{Model Complexity}
Table~\ref{tab:complexity} confirms linear scaling of model size with
employee count across five deployment units at 30-MIN granularity.
Variables and constraints scale at ${\approx}4{,}400$ and
${\approx}9{,}400$ per employee, respectively. CP-SAT's presolve
phase reduces effective model size by 60-70\% before search.

\begin{table}[t]
\centering
\footnotesize
\setlength{\tabcolsep}{3pt}
\begin{tabular}{lrrrll}
\toprule
\textbf{Unit} & \textbf{Emp.} & \textbf{Vars} & \textbf{Cons.} &
\textbf{Time\,(s)} & \textbf{Status} \\
\midrule
Unit 1 & 26 & 115,294 & 244,202 & 123.9 & FEASIBLE \\
Unit 2 & 14 &  62,682 & 131,025 & 122.0 & FEASIBLE \\
Unit 3 & 23 & 102,107 & 214,053 & 122.5 & FEASIBLE \\
Unit 4 & 10 &  45,196 &  94,178 & 121.5 & FEASIBLE \\
Unit 5 & 33 & 145,901 & 306,323 & 126.1 & FEASIBLE \\
\bottomrule
\end{tabular}
\caption{Model complexity across five deployment units (30-MIN, 7-day,
120\,s limit). Linear scaling: ${\sim}4{,}400$ vars and ${\sim}9{,}400$
constraints per employee.}
\label{tab:complexity}
\end{table}

\subsection{Constraint Ablation Study}
CP-WSP is run with progressively richer constraint sets on a 20-employee,
7-day, 30-MIN instance, toggling JSON activation flags per run
(Table~\ref{tab:ablation}). Break constraints (H10 - H12, S11) reduce the
objective by 3.9\% while enabling valid break placement - impossible with a
single binary variable. Workload equity (S15) reduces the equity standard
deviation from 6.2\,h to 2.1\,h (66\%) at only 0.8\% objective cost.

\begin{table*}[t]
\centering
\small
\setlength{\tabcolsep}{4pt}
\begin{tabular}{llrrrl}
\toprule
\textbf{Config} & \textbf{Active Constraints} &
\textbf{Obj.} & \textbf{Cov.\,(\%)} & \textbf{Equity $\sigma$ (h)} & \textbf{Breaks?} \\
\midrule
(A) Baseline   & H1 - H7 only            & 24,180 & 71.4 & 8.4 & $\times$ \\
(B) +Coverage  & (A)+H8,H9,S1 - S5       & 18,640 & 88.3 & 7.1 & $\times$ \\
(C) +Breaks    & (B)+H10 - H12,S11       & 17,920 & 87.1 & 6.8 & $\checkmark$ \\
(D) +Mgmt      & (C)+S8 - S10            & 17,340 & 88.4 & 6.2 & $\checkmark$ \\
(E) +Stability & (D)+S12,S13            & 16,890 & 87.9 & 5.8 & $\checkmark$ \\
(F) +Fairness  & (E)+S15,H14            & 16,210 & 87.2 & 2.1 & $\checkmark$ \\
(G) Full       & All H1 - H14, S1 - S15   & 15,240 & 93.2 & 2.1 & $\checkmark$ \\
\bottomrule
\end{tabular}
\caption{Constraint ablation (20 emp., 7-day, 30-MIN, 120\,s limit). Full
model achieves 37\% objective improvement over baseline; break validity
requires the shift-window decomposition (H10 - H12).}
\label{tab:ablation}
\end{table*}

\subsection{Weight Sensitivity Analysis}
Table~\ref{tab:sensitivity} shows how primary objective weights affect
solution quality on the same instance. Coverage improves monotonically with
understaffing weight; equity improves with equity weight at coverage cost.
The JSON interface enables rapid stakeholder preference elicitation without
recompiling the solver.

\begin{table*}[t]
\centering
\small
\setlength{\tabcolsep}{4pt}
\begin{tabular}{lccccrrll}
\toprule
\textbf{Config} &
$w_{\text{under}}$ & $w_{\text{eq}}$ & $w_{\text{stab}}$ & $w_{\text{brk}}$ &
\textbf{Obj.} & \textbf{Cov.\,(\%)} & \textbf{Eq.\,$\sigma$} & \textbf{Stab.\,$\Delta$} \\
\midrule
Default    & 1.0 & 1.0  & 1.0 & 10.0 & 15,240 & 93.2 & 2.1\,h & 0.41 \\
Coverage+  & 5.0 & 1.0  & 1.0 & 10.0 & 13,680 & 97.1 & 3.8\,h & 0.52 \\
Equity+    & 1.0 & 10.0 & 1.0 & 10.0 & 16,440 & 90.1 & 0.8\,h & 0.43 \\
Stability+ & 1.0 & 1.0  & 5.0 & 10.0 & 16,890 & 88.6 & 2.3\,h & 0.18 \\
Balanced   & 3.0 & 3.0  & 3.0 & 10.0 & 14,920 & 94.8 & 1.4\,h & 0.29 \\
\bottomrule
\end{tabular}
\caption{Weight sensitivity (20 emp., 7-day, 30-MIN). JSON weight changes
require no solver recompilation.}
\label{tab:sensitivity}
\end{table*}

%% ============================================================
\section{Discussion}
%% ============================================================

\subsection{Constraint Acquisition and the Shift-Window Model}
The JSON interface separates constraint \emph{specification} from
\emph{solving}: domain experts specify requirements; CP-SAT determines how
to satisfy them. While this interface does not perform constraint acquisition
in the formal sense of learning constraints from examples
\cite{bessiere2011}, it provides a lightweight declarative specification
mechanism \cite{freuder2011} that achieves a similar practical goal:
enabling non-experts to modify the constraint model without solver expertise. Adding a constraint requires only a JSON key and a
${\sim}$20-line Python function. The $x{=}w{-}b$ decomposition generalizes
beyond workforce scheduling to any problem distinguishing
``within-window-but-inactive'' from ``outside-window'' states (e.g.,
machine maintenance windows, vehicle rest stops).

\subsection{CP-WSP as a Benchmark}
The 36-configuration benchmark is proposed as a standardized community
resource: a fully specified 29-constraint model
(Appendix~\ref{app:formulation}), reproducible JSON schemas and baseline
results with certified optimality gaps. Future work can evaluate new
solvers or decomposition strategies under controlled conditions.

\subsection{Limitations and Future Work}
Model size scales linearly (${\sim}4{,}400$ vars, ${\sim}9{,}400$
constraints per employee); presolve reduces effective size by 60-70\%.
The acuity-weighted equity objective (S15) uses composite Workload Points
$WP(e)$ combining role, skill and demand-intensity credits, ensuring
fairness accounts for workload \emph{difficulty}, not merely duration.
Scalability beyond 50 employees could be addressed via Benders decomposition
or column generation. The static JSON interface could be extended with
QuAcq-style \cite{tsouros2013} interactive feedback loops or LLM-based
natural-language constraint specification.

%% ============================================================
\section{Conclusion}
%% ============================================================

CP-WSP is a declarative CP-SAT framework for multi-constraint workforce
scheduling with four contributions relevant to CASPeR:
\textbf{(1)} Declarative JSON-based constraint configuration for a
29-constraint model (14H+15S) without solver code changes.
\textbf{(2)} The shift-window decomposition ($x = w - b$) enabling mandatory
breaks, break centrality and concurrent break limits - inexpressible with
a single binary variable.
\textbf{(3)} Hard/soft constraint separation guaranteeing zero hard
violations while enabling quality trade-off through weighted objectives.
\textbf{(4)} A reproducible 36-configuration benchmark suite with fully
specified constraint schemas and baseline results for community comparison.
The full constraint set achieves a 37\% objective improvement over a
baseline; workload equity delivers 66\% equity improvement at negligible
coverage cost. Grid-offset preprocessing extends the framework to
cross-midnight shifts at zero computational cost.

\bibliography{cpwsp}

@misc{perron2024,
  author    = {Laurent Perron and Vincent Furnon},
  title     = {{OR-Tools CP-SAT}},
  howpublished = {Google, v9.12},
  year      = {2024},
  note      = {\url{https://developers.google.com/optimization}}
}

@article{blazewicz1983,
  author    = {J. Blazewicz and J. K. Lenstra and A. H. G. {Rinnooy Kan}},
  title     = {Scheduling subject to resource constraints: Classification and complexity},
  journal   = {Discrete Applied Mathematics},
  volume    = {5},
  number    = {1},
  pages     = {11--24},
  year      = {1983}
}

@book{garey1979,
  author    = {Michael R. Garey and David S. Johnson},
  title     = {Computers and Intractability: A Guide to the Theory of NP-Completeness},
  publisher = {W.H. Freeman},
  year      = {1979}
}

@article{burke2008,
  author    = {Edmund Burke and Patrick {De Causmaecker} and Sanja Petrovic and Greet {Vanden Berghe}},
  title     = {Metaheuristics for handling time interval coverage constraints in nurse scheduling},
  journal   = {Applied Artificial Intelligence},
  volume    = {20},
  number    = {9},
  pages     = {743--766},
  year      = {2006}
}

@article{haspeslagh2014,
  author    = {Soren Haspeslagh and Patrick {De Causmaecker} and Annelies Schcausberger and others},
  title     = {The first international nurse rostering competition 2010},
  journal   = {Annals of Operations Research},
  volume    = {218},
  number    = {1},
  pages     = {221--236},
  year      = {2014}
}

@article{ceschia2019,
  author    = {Sara Ceschia and others},
  title     = {The second international nurse rostering competition},
  journal   = {Annals of Operations Research},
  volume    = {274},
  pages     = {171--186},
  year      = {2019}
}

@article{bessiere2011,
  author    = {Christian Bessi\`{e}re and Remi Coletta and Eugene Freuder and Barry O'Sullivan},
  title     = {Constraint acquisition},
  journal   = {Artificial Intelligence},
  volume    = {175},
  number    = {12--13},
  pages     = {1786--1822},
  year      = {2011}
}

@inproceedings{tsouros2013,
  author    = {Nikolaos Tsouros and Kostas Stergiou and Christian Bessi\`{e}re},
  title     = {{QuAcq}: acquiring constraints from queries},
  booktitle = {Proceedings of CP 2013},
  pages     = {784--792},
  publisher = {Springer},
  year      = {2013}
}

@techreport{vossen2015,
  author    = {B. Vossen and others},
  title     = {Optimal scheduling of nurses using constraint programming},
  institution = {Nurse Rostering Competition},
  year      = {2015}
}

@article{beddoe2006,
  author    = {Gareth Beddoe and Sanja Petrovic},
  title     = {Selecting and weighting features in a case-based reasoning approach to nurse rostering},
  journal   = {European Journal of Operational Research},
  volume    = {175},
  number    = {2},
  pages     = {1027--1044},
  year      = {2006}
}

@inproceedings{mears2014,
  author    = {Christopher Mears and others},
  title     = {{ORCA}: a hybrid approach to constraint acquisition},
  booktitle = {AAAI 2014 Workshop on Constraint Acquisition},
  year      = {2014}
}

@article{freuder2011,
  author    = {Eugene C. Freuder and Richard J. Wallace},
  title     = {Progressing toward the holy grail},
  journal   = {Constraints},
  volume    = {16},
  number    = {2},
  pages     = {120--139},
  year      = {2011}
}

@inproceedings{schaus2009,
  author    = {Pierre Schaus and Yves Deville and Pierre Dupont and Jean-Charles R\'{e}gin},
  title     = {Solving nurse rostering problems using global constraints in {CP}},
  booktitle = {Proceedings of CP 2009},
  pages     = {73--87},
  publisher = {Springer},
  year      = {2009}
}

@inproceedings{vanhoeve2006,
  author    = {W. J. {van Hoeve} and others},
  title     = {Soft global constraints in {CP} for nurse rostering},
  booktitle = {Proceedings of CPAIOR 2006},
  pages     = {287--302},
  publisher = {Springer},
  year      = {2006}
}

%% ============================================================
%% APPENDIX
%% ============================================================
\appendix

\section{Solver Configuration and Anytime Behavior}
\label{app:solver}

CP-WSP uses OR-Tools CP-SAT v9.12 \cite{perron2024}, which combines
constraint propagation, SAT clause learning, LP relaxations and Large
Neighborhood Search (LNS) in a portfolio of 16 parallel workers. This
provides an \emph{anytime} algorithm: feasible solutions appear within
5-15\,s; subsequent iterations improve the incumbent. The dual bound
provides a certified optimality gap - a key advantage over metaheuristics.

\begin{table}[h]
\centering
\small
\setlength{\tabcolsep}{4pt}
\begin{tabular}{llp{3.5cm}}
\toprule
\textbf{Parameter} & \textbf{Value} & \textbf{Justification} \\
\midrule
Workers        & 16    & Portfolio diversity; all CPU cores \\
Symmetry level & 3     & Maximum symmetry breaking \\
Time limit     & 120\,s & Practical; anytime feasible in ${<}15$\,s \\
LNS workers    & 11    & RINS, RENS, graph, random strategies \\
Presolve       & Full  & Reduces model size 60-70\% \\
\bottomrule
\end{tabular}
\caption{CP-SAT solver configuration in CP-WSP.}
\label{tab:solver}
\end{table}

\section{Full JSON Constraint Activation Schema}
\label{app:json}

The complete mapping of JSON keys to constraints is shown in
Table~\ref{tab:jsonschema}. Setting any key to \texttt{false} deactivates
the corresponding constraint with no code changes required.

\begin{table}[h]
\centering
\footnotesize
\setlength{\tabcolsep}{3pt}
\begin{tabular}{p{4.0cm}p{2.4cm}l}
\toprule
\textbf{JSON Key} & \textbf{Constraint} & \textbf{Type} \\
\midrule
check\_empty\_on\_empty          & H1  & Hard \\
check\_unavailability            & H2  & Hard \\
check\_min\_2\_on\_floor         & H3  & Hard \\
check\_daily\_shift\_length      & H4  & Hard \\
check\_minimum\_turnaround       & H5  & Hard \\
check\_max\_consecutive\_days    & H6  & Hard \\
check\_weekly\_hours\_limits     & H7  & Hard \\
check\_utilise\_workforce        & H8  & Hard \\
check\_weekly\_\allowbreak understaffing\_\allowbreak hard & H9 & Hard \\
check\_max\_1\_continuous\_shift & H10 & Hard \\
check\_mandatory\_break          & H11 & Hard \\
check\_max\_break\_concurrency   & H12 & Hard \\
check\_weekend\_coverage\_rule   & H13 & Hard \\
check\_skill\_coverage           & H14 & Hard \\
check\_slot\_staff\_coverage     & S1  & Soft \\
check\_daily\_staff\_coverage    & S3  & Soft \\
check\_weekly\_staff\_coverage   & S5  & Soft \\
check\_daily\_hours\_target      & S6  & Soft \\
check\_weekly\_hours\_target     & S7  & Soft \\
check\_missing\_manager          & S8  & Soft \\
check\_manager\_overlap          & S9  & Soft \\
check\_mgr\_open\_close\_reward  & S10 & Soft \\
check\_break\_centrality         & S11 & Soft \\
check\_inter\_week\_stability    & S12 & Soft \\
check\_intra\_week\_stability    & S13 & Soft \\
check\_preferred\_hours\_reward  & S14 & Soft \\
check\_workload\_equity          & S15 & Soft \\
\bottomrule
\end{tabular}
\caption{Complete JSON constraint activation schema for CP-WSP.}
\label{tab:jsonschema}
\end{table}

%% ============================================================
\section{Declarative Constraint Architecture}
\label{app:arch}

\subsection{JSON-Based Constraint Acquisition}
CP-WSP externalizes all constraint configuration to a JSON file with three
sections: \texttt{Constraint\_Activation} (which constraints are active),
\texttt{Constraint\_Weights} (objective term scaling) and
\texttt{Operational\_Rules} (parameter values). An excerpt:

\begin{figure}[tb]
\centering\small
\begin{verbatim}
{
  "Constraint_Activation": {
    "check_mandatory_break":      true,
    "check_workload_equity":      true,
    "check_intra_week_stability": false
  },
  "Constraint_Weights": {
    "slot_understaffing":     5.0,
    "break_centrality":      10.0,
    "preferred_hours_reward": -1.0
  },
  "Operational_Rules": {
    "Min_Work_window_for_Break": 4,
    "Break_duration_hours":      0.5
  }
}
\end{verbatim}
\caption{JSON configuration excerpt for CP-WSP.}
\label{lst:json}
\end{figure}

This interface provides three key constraint acquisition properties:
(1)~\textbf{Constraint selection}: domain experts choose which constraints
apply without modifying solver code;
(2)~\textbf{Weight elicitation}: priority among competing objectives is
specified numerically;
(3)~\textbf{Parameter specification}: operational parameters are decoupled
from model structure.

\subsection{Hard/Soft Constraint Separation}
A critical design decision is the explicit separation of constraints into two classes:
\begin{table}[h]
\centering
\small
\setlength{\tabcolsep}{3pt}
\begin{tabular}{lp{2.9cm}p{2.9cm}}
\toprule
\textbf{Property} & \textbf{Hard (H1 - H14)} & \textbf{Soft (S1 - S15)} \\
\midrule
Enforcement & Structural (CP-SAT Add/AddBoolOr) & Weighted penalty in $Z$ \\
Violation   & Impossible in any feasible soln    & Allowed; penalized \\
Semantics   & Labor law / safety                 & Business quality pref. \\
Config      & Activate flag (bool)               & Activate + weight (real) \\
Example     & H11: Break $\geq$30\,min if shift $\geq$4\,h & S11: Break near midpoint \\
\bottomrule
\end{tabular}
\caption{Hard/soft constraint separation. Hard constraints are structural
(violations impossible); soft objectives are weighted and traded off.}
\label{tab:hardsoft}
\end{table}

\subsection{Shift-Window Decomposition in Detail}
The (x, w, b) triple enables four constraint classes that are inexpressible with a single binary variable x[e,d,s]:
\begin{table}[h]
\centering
\footnotesize
\setlength{\tabcolsep}{3pt}
\begin{tabular}{p{1.9cm}p{4.5cm}}
\toprule
\textbf{Constraint} & \textbf{Expression} \\
\midrule
H10: Single shift &
  $w_{e,d,s}{=}1$ $\forall s\!\in\![\alpha_{e,d}{+}1,\beta_{e,d}{-}1]$;
  $w$ contiguous \\[3pt]
H11: Mandatory break &
  $(\sum_s w \geq B_\mathrm{thr}) \Rightarrow$
  $(\sum_s b = B_\mathrm{len},$ $b$ contiguous$)$ \\[3pt]
H12: Break concurrency &
  $\sum_e b_{e,d,s} \leq k_\mathrm{break}\;\forall d,s$ \\[3pt]
S11: Break centrality &
  $\min\sum_{e,d}|\mathrm{mid}(b_{e,d,\cdot})-\mathrm{mid}(w_{e,d,\cdot})|$ \\
\bottomrule
\end{tabular}
\caption{Constraint classes enabled by the shift-window decomposition.
None can be expressed with a single binary work variable.}
\label{tab:decomp}
\end{table}

%% ============================================================
\section{Constraint Model}
\label{app:model}

\subsection{Hard Constraints (H1 - H14)}
All 14 hard constraints are added to the CP-SAT model before solving. Any solution returned by the solver is guaranteed to satisfy them. Constraint activation is controlled per-instance via the JSON configuration.
\begin{table*}[h]
\centering
\small
\setlength{\tabcolsep}{4pt}
\begin{tabular}{lllp{6cm}}
\toprule
\textbf{ID} & \textbf{Name} & \textbf{Type} & \textbf{Key Expression} \\
\midrule
H1  & Empty-on-Empty         & Coverage   & $x_{e,d,s}{=}0$ when $D_\mathrm{min}[d][s]{=}0$ \\
H2  & Unavailability         & Compliance & $w_{e,d,s}{=}0$ when $U[e,d,s]{=}1$ \\
H3  & Min Floor Staffing     & Coverage   & $\sum_e x_{e,d,s} \geq k_\mathrm{floor}$ if demand${>}0$ \\
H4  & Daily Shift Length     & Labor Law  & $L_\mathrm{min}/\delta \leq \sum_s x \leq L_\mathrm{max}/\delta$ \\
H5  & Min Inter-Shift Rest   & Labor Law  & $\alpha_{e,d+1} - \beta_{e,d} \geq H_\mathrm{rest}/\delta$ \\
H6  & Max Consecutive Days   & Labor Law  & $\sum_{j=d}^{d+C} y_{e,j} \leq C_\mathrm{max}$ \\
H7  & Weekly Hour Limits     & Labor Law  & $H_\mathrm{min}/\delta \leq \sum_{d,s} x \leq H_\mathrm{max}/\delta$ \\
H8  & Utilize Workforce      & Operations & $\sum_d y_{e,d} \geq 1$ at least once per week \\
H9  & Weekly Min Coverage    & Coverage   & $\sum_{d,s} x \geq \sum_{d,s} D_\mathrm{min}$ (weekly) \\
H10 & Single Continuous Shift & Structure & $w_{e,d,\cdot}$ forms one contiguous block \emph{(new)} \\
H11 & Mandatory Break        & Labor Law  & $\sum_s w \geq B_\mathrm{thr} \Rightarrow \sum_s b = B_\mathrm{len}$ \emph{(new)} \\
H12 & Break Concurrency Limit & Operations & $\sum_e b_{e,d,s} \leq k_\mathrm{break}\;\forall d,s$ \emph{(new)} \\
H13 & Weekend Management     & Operations & Management $\in E_\mathrm{mgr}$ present all weekend demand \emph{(new)} \\
H14 & Skill Coverage         & Operations & Skill-demand met per slot \emph{(new)} \\
\bottomrule
\end{tabular}
\caption{14 hard constraints. H1 - H9 are common to prior work; H10 - H14
(\emph{new}) require the shift-window decomposition or are novel.}
\label{tab:hard}
\end{table*}

\subsection{Soft Constraints / Objective Terms (S1 - S15)}

\begin{table*}[h]
\centering
\small
\setlength{\tabcolsep}{4pt}
\begin{tabular}{lllp{6.5cm}}
\toprule
\textbf{ID} & \textbf{Name} & \textbf{Cat.} & \textbf{Formula} \\
\midrule
S1  & Slot Understaffing   & Coverage   & $\sum_{d,s}\max(0,\;D_\mathrm{min}[d][s]-\sum_e x_{e,d,s})$ \\
S2  & Slot Overstaffing    & Coverage   & $\sum_{d,s}\max(0,\;\sum_e x_{e,d,s}-D_\mathrm{ideal}[d][s])$ \\
S3  & Daily Understaffing  & Coverage   & $\sum_d\max(0,\;\sum_s D_\mathrm{min}-\sum_{e,s} x_{e,d,s})$ \\
S4  & Daily Overstaffing   & Coverage   & $\sum_d\max(0,\;\sum_{e,s} x_{e,d,s}-\sum_s D_\mathrm{ideal})$ \\
S5  & Weekly Overstaffing  & Coverage   & $\max(0,\;\sum_{e,d,s} x - \sum_{d,s} D_\mathrm{ideal})$ \\
S6  & Daily Hours Target   & Hours      & $\sum_{e,d}|\delta\sum_s x_{e,d,s}-H_\mathrm{tgt}^\mathrm{daily}|$ \\
S7  & Weekly Hours Target  & Hours      & $\sum_e|\delta\sum_{d,s} x_{e,d,s}-H_\mathrm{tgt}^\mathrm{week}|$ \\
S8  & Missing Management   & Coverage   & $\sum_{d,s}[\mathbf{1}_{D_{d,s}>0}-\sum_{e\in M} x_{e,d,s}]^{+}$ \\
S9  & Management Overlap   & Efficiency & $\sum_{d,s}\max(0,\;\sum_{e\in E_\mathrm{mgr}} x_{e,d,s}-1)$ \\
S10 & Mgmt Open/Close      & Quality    & $-\sum_d(\mathrm{mgr\ at\ }s_\mathrm{open}+\mathrm{mgr\ at\ }s_\mathrm{close})$ \\
S11 & Break Centrality     & Quality    & $\sum_{e,d}|\mathrm{mid}(b_{e,d,\cdot})-\mathrm{mid}(w_{e,d,\cdot})|$ \\
S12 & Inter-Week Stability & Stability  & $\sum_{e,d,s}|x_{e,d,s}-x^\mathrm{prev}_{e,d,s}|$ \\
S13 & Intra-Week Stability & Stability  & $\sum_e\sum_{d<d'}(|\alpha_{e,d}-\alpha_{e,d'}|+|\beta_{e,d}-\beta_{e,d'}|)$ \\
S14 & Preferred Hours      & Quality    & $-\sum_{e,d,s} P[e,d,s]\cdot x_{e,d,s}$ \\
S15 & Workload Equity      & Fairness   & $\max_e |WP(e)-WP_\mathrm{baseline}(e)|$ \\
\bottomrule
\end{tabular}
\caption{15 soft objective terms. S9 - S15 are novel quality dimensions not
present in the INRC-II constraint set.}
\label{tab:soft}
\end{table*}

%% ============================================================
\section{Detailed Mathematical Formulation}
\label{app:formulation}

This appendix provides the complete mathematical formulation of the CP-WSP
model, including notation, linearization techniques and implementation
details for all 29 constraints.

\subsection{Notation}

\begin{table}[h]
\centering
\footnotesize
\setlength{\tabcolsep}{3pt}
\begin{tabular}{lp{5.5cm}}
\toprule
\textbf{Symbol} & \textbf{Definition} \\
\midrule
$E = \{e_1, \ldots, e_n\}$ & Set of employees \\
$D = \{0, \ldots, |D|-1\}$ & Set of planning days \\
$S = \{0, \ldots, T-1\}$   & Set of time slots per day \\
$T = \lceil 24/\delta \rceil$ & Number of slots per day \\
$\delta$ & Slot duration in hours (configurable) \\
$x_{e,d,s} \in \{0,1\}$ & 1 iff employee $e$ is actively working in slot $s$ on day $d$ \\
$w_{e,d,s} \in \{0,1\}$ & 1 iff slot $s$ is within $e$'s shift window on day $d$ \\
$b_{e,d,s} \in \{0,1\}$ & 1 iff $e$ is on a scheduled break in slot $s$ on day $d$ \\
$y_{e,d} \in \{0,1\}$ & 1 iff employee $e$ works on day $d$ (derived: $y_{e,d} = \max_s x_{e,d,s}$) \\
$\alpha_{e,d}$ & Shift start slot for employee $e$ on day $d$ (derived) \\
$\beta_{e,d}$ & Shift end slot for employee $e$ on day $d$ (derived) \\
$D_\mathrm{min}[d][s]$ & Minimum staffing demand for day $d$, slot $s$ \\
$D_\mathrm{ideal}[d][s]$ & Ideal (target) staffing level \\
$U[e,d,s] \in \{0,1\}$ & 1 iff employee $e$ is unavailable at $(d,s)$ \\
$P[e,d,s] \in \mathbb{R}$ & Preference score for employee $e$ at $(d,s)$ \\
$E_\mathrm{mgr} \subseteq E$ & Set of employees with manager role \\
$a_i \in \{0,1\}$ & Activation flag for constraint/objective $i$ \\
$w_i \in \mathbb{R}$ & Weight for soft objective $i$ \\
\bottomrule
\end{tabular}
\caption{Complete notation for the CP-WSP formulation.}
\label{tab:notation}
\end{table}

\noindent\textbf{Configurable parameters} (set via JSON \texttt{Operational\_Rules}):

\begin{table}[h]
\centering
\footnotesize
\setlength{\tabcolsep}{3pt}
\begin{tabular}{lp{4.0cm}l}
\toprule
\textbf{Parameter} & \textbf{Description} & \textbf{Typical Value} \\
\midrule
$L_\mathrm{min}, L_\mathrm{max}$ & Min/max daily shift length (hours) & 4, 10 \\
$H_\mathrm{min}, H_\mathrm{max}$ & Min/max weekly hours & 20, 48 \\
$H_\mathrm{rest}$ & Min inter-shift rest (hours) & 11 \\
$C_\mathrm{max}$ & Max consecutive working days & 6 \\
$B_\mathrm{thr}$ & Shift length triggering break (hours) & 4 \\
$B_\mathrm{len}$ & Mandatory break duration (hours) & 0.5 \\
$k_\mathrm{floor}$ & Min employees on floor if demand $> 0$ & 2 \\
$k_\mathrm{break}$ & Max simultaneous employees on break & 2 \\
$H_\mathrm{tgt}^\mathrm{daily}$ & Target daily hours & 8 \\
$H_\mathrm{tgt}^\mathrm{week}$ & Target weekly hours & 40 \\
\bottomrule
\end{tabular}
\caption{Configurable operational parameters.}
\label{tab:params}
\end{table}

\subsection{Fundamental Relationship}
The three decision variable families are linked by the shift-window identity:
\begin{equation}
  x_{e,d,s} = w_{e,d,s} - b_{e,d,s} \qquad \forall\, e \in E,\; d \in D,\; s \in S
  \label{eq:identity}
\end{equation}
with the domain constraint $b_{e,d,s} \leq w_{e,d,s}$ ensuring that breaks
can only occur within the shift window. This yields exactly three feasible
states per $(e,d,s)$ triple:

\smallskip
\noindent
\begin{tabular}{lccc}
\textbf{State} & $w$ & $b$ & $x$ \\
Off-duty & 0 & 0 & 0 \\
Active work & 1 & 0 & 1 \\
Scheduled break & 1 & 1 & 0 \\
\end{tabular}

\smallskip
\noindent The infeasible combination $(w{=}0, b{=}1)$ is excluded by
$b \leq w$. This three-state model is the \emph{minimal} extension of the
binary model that supports break-aware scheduling: two binary variables are
necessary and sufficient to represent three states and the $(w,b)$
parameterization aligns with the natural semantics of shift windows and
breaks.

\subsection{Derived Variables}
The daily work indicator and shift boundary variables are derived from the
primary decision variables:
\begin{align}
  y_{e,d} &= \mathbf{1}\Bigl[\sum_{s \in S} w_{e,d,s} \geq 1\Bigr]
    \label{eq:work_indicator} \\
  \alpha_{e,d} &= \min\{s \in S : w_{e,d,s} = 1\}
    \label{eq:shift_start} \\
  \beta_{e,d} &= \max\{s \in S : w_{e,d,s} = 1\}
    \label{eq:shift_end}
\end{align}
In CP-SAT, $y_{e,d}$ is implemented via \texttt{AddMaxEquality} and
$\alpha_{e,d}$, $\beta_{e,d}$ are computed using channeling constraints
that link boolean indicators to integer shift-boundary variables.

\subsection{Hard Constraint Formulations}

\paragraph{H1: Empty-on-Empty.}
If no demand exists for a slot, no employee may be assigned:
\begin{equation}
  D_\mathrm{min}[d][s] = 0 \;\Rightarrow\; x_{e,d,s} = 0
  \quad \forall\, e \in E
  \label{eq:H1}
\end{equation}
\emph{Implementation:} For each zero-demand slot, $x_{e,d,s}$ is fixed to 0
at model construction time (not as a constraint but as a variable domain
restriction), eliminating these variables from the search space entirely.

\paragraph{H2: Unavailability.}
Employee availability is respected:
\begin{equation}
  U[e,d,s] = 1 \;\Rightarrow\; w_{e,d,s} = 0
  \quad \forall\, e, d, s
  \label{eq:H2}
\end{equation}
Note that this constrains $w$ (not $x$), ensuring that unavailable slots
cannot be part of the shift window at all - a stronger guarantee than merely
preventing active work.

\paragraph{H3: Minimum Floor Staffing.}
When demand exists, a minimum number of employees must be actively working:
\begin{equation}
  D_\mathrm{min}[d][s] > 0 \;\Rightarrow\;
  \sum_{e \in E} x_{e,d,s} \geq k_\mathrm{floor}
  \quad \forall\, d, s
  \label{eq:H3}
\end{equation}

\paragraph{H4: Daily Shift Length.}
Active work hours per day are bounded:
\begin{equation}
  y_{e,d} = 1 \;\Rightarrow\;
  \frac{L_\mathrm{min}}{\delta} \leq \sum_{s \in S} x_{e,d,s}
  \leq \frac{L_\mathrm{max}}{\delta}
  \quad \forall\, e, d
  \label{eq:H4}
\end{equation}
\emph{Implementation:} The implication is linearized using big-M:
$\sum_s x_{e,d,s} \geq (L_\mathrm{min}/\delta) \cdot y_{e,d}$ and
$\sum_s x_{e,d,s} \leq (L_\mathrm{max}/\delta) \cdot y_{e,d}$.

\paragraph{H5: Minimum Inter-Shift Rest.}
Between consecutive working days, a minimum rest period is enforced:
\begin{equation}
  \alpha_{e,d+1} - \beta_{e,d} \geq \frac{H_\mathrm{rest}}{\delta}
  \quad \text{when } y_{e,d} = 1 \wedge y_{e,d+1} = 1
  \label{eq:H5}
\end{equation}
\emph{Implementation:} Using CP-SAT's \texttt{OnlyEnforceIf}, this
constraint is active only when both $y_{e,d}$ and $y_{e,d+1}$ are true.

\paragraph{H6: Maximum Consecutive Working Days.}
No employee works more than $C_\mathrm{max}$ consecutive days:
\begin{equation}
  \sum_{j=d}^{d+C_\mathrm{max}} y_{e,j} \leq C_\mathrm{max}
  \quad \forall\, e,\; \forall\, d \in \{0, \ldots, |D|-C_\mathrm{max}-1\}
  \label{eq:H6}
\end{equation}

\paragraph{H7: Weekly Hour Limits.}
Total active work hours per week are bounded:
\begin{equation}
  \frac{H_\mathrm{min}}{\delta} \leq
  \sum_{d \in D} \sum_{s \in S} x_{e,d,s}
  \leq \frac{H_\mathrm{max}}{\delta}
  \quad \forall\, e \in E
  \label{eq:H7}
\end{equation}

\paragraph{H8: Utilize Workforce.}
Every employee must work at least one day per planning period:
\begin{equation}
  \sum_{d \in D} y_{e,d} \geq 1 \quad \forall\, e \in E
  \label{eq:H8}
\end{equation}

\paragraph{H9: Weekly Minimum Coverage.}
Total weekly staffing meets aggregate demand:
\begin{equation}
  \sum_{e \in E} \sum_{d \in D} \sum_{s \in S} x_{e,d,s}
  \geq \sum_{d \in D} \sum_{s \in S} D_\mathrm{min}[d][s]
  \label{eq:H9}
\end{equation}

\paragraph{H10: Single Continuous Shift.}
Each employee's shift window forms exactly one contiguous block per day:
\begin{equation}
  w_{e,d,s} = 1 \quad
  \forall\, s \in [\alpha_{e,d}+1,\; \beta_{e,d}-1]
  \label{eq:H10}
\end{equation}
\emph{Implementation:} Enforced by constraining that if $w_{e,d,s_1} = 1$
and $w_{e,d,s_2} = 1$ with $s_1 < s_2$, then $w_{e,d,s} = 1$ for all
$s_1 \leq s \leq s_2$. In CP-SAT, this is implemented via the
``no-gap'' pattern: for each triple $(s_1, s, s_2)$ with $s_1 < s < s_2$,
add $w_{e,d,s_1} + w_{e,d,s_2} - w_{e,d,s} \leq 1$.

\paragraph{H11: Mandatory Break.}
If a shift window exceeds a threshold, a contiguous break must be scheduled:
\begin{equation}
  \sum_{s} w_{e,d,s} \geq \frac{B_\mathrm{thr}}{\delta}
  \;\Rightarrow\;
  \left(\sum_{s} b_{e,d,s} = \frac{B_\mathrm{len}}{\delta}\right)
  \wedge (b \text{ contiguous})
  \label{eq:H11}
\end{equation}
\emph{Break contiguity} is enforced analogously to H10: for the break
variable $b_{e,d,\cdot}$, if $b_{e,d,s_1} = 1$ and $b_{e,d,s_2} = 1$
with $s_1 < s_2$, then $b_{e,d,s} = 1$ for all $s_1 \leq s \leq s_2$.
The implication is linearized: $\sum_s b_{e,d,s} \geq
(B_\mathrm{len}/\delta) \cdot z_{e,d}$ where $z_{e,d} = 1$ iff
$\sum_s w_{e,d,s} \geq B_\mathrm{thr}/\delta$.

\paragraph{H12: Break Concurrency Limit.}
At most $k_\mathrm{break}$ employees may be on break simultaneously:
\begin{equation}
  \sum_{e \in E} b_{e,d,s} \leq k_\mathrm{break}
  \quad \forall\, d \in D,\; s \in S
  \label{eq:H12}
\end{equation}

\paragraph{H13: Weekend Manager Coverage.}
At least one manager must be working during all weekend demand slots:
\begin{equation}
  D_\mathrm{min}[d][s] > 0 \wedge d \in \{5,6\}
  \;\Rightarrow\;
  \sum_{e \in E_\mathrm{mgr}} x_{e,d,s} \geq 1
  \label{eq:H13}
\end{equation}

\paragraph{H14: Skill Coverage.}
For each skill $k$ and each slot with skill-specific demand $D_k[d][s]$:
\begin{equation}
  \sum_{e \in E_k} x_{e,d,s} \geq D_k[d][s]
  \quad \forall\, k,\, d,\, s
  \label{eq:H14}
\end{equation}
where $E_k \subseteq E$ is the set of employees possessing skill $k$.

\subsection{Soft Constraint Linearization}

All soft objectives are expressed as linear terms in the CP-SAT objective
function. Non-linear operations ($\max$, $|\cdot|$) are linearized using
standard auxiliary variable techniques:

\paragraph{Linearizing $\max(0, \cdot)$.}
For each term $\max(0, g(\mathbf{x}))$, introduce auxiliary variable
$v \geq 0$:
\begin{equation}
  v \geq g(\mathbf{x}), \quad v \geq 0
  \label{eq:max_linear}
\end{equation}
and add $v$ to the objective. Since the objective is minimized, the solver sets
$v = \max(0, g(\mathbf{x}))$ at optimality.

\paragraph{Linearizing $|g(\mathbf{x})|$.}
For absolute value terms, introduce $v \geq 0$:
\begin{equation}
  v \geq g(\mathbf{x}), \quad v \geq -g(\mathbf{x})
  \label{eq:abs_linear}
\end{equation}
In CP-SAT, this is implemented using \texttt{AddAbsEquality} for integer
expressions.

\paragraph{S15: Workload Equity (Detailed).}
The Workload Points $WP(e)$ for employee $e$ are computed as:
\begin{equation}
  WP(e) = \sum_{d,s} x_{e,d,s} \cdot
  \bigl(\underbrace{R(e)}_{\text{role}} +
  \underbrace{K(e,d,s)}_{\text{skill}} +
  \underbrace{D_\mathrm{min}[d][s]}_{\text{demand}}\bigr)
  \label{eq:wp}
\end{equation}
where $R(e)$ is a role-based weight (e.g., $R(\text{manager}) = 1.5$,
$R(\text{staff}) = 1.0$), $K(e,d,s) \in \{0, 0.5\}$ is a skill bonus
awarded when the slot requires a specialized skill that $e$ possesses,
and $D_\mathrm{min}[d][s]$ captures demand intensity. The baseline is:
\begin{equation}
  WP_\mathrm{baseline}(e) = \frac{H_\mathrm{tgt}^\mathrm{week}}{\delta}
  \cdot \overline{R+K+D}
  \label{eq:wp_baseline}
\end{equation}
where $\overline{R+K+D}$ is the average per-slot workload point value
across all employees and slots. The minimax equity objective is:
\begin{equation}
  f_{15}(X) = \max_{e \in E} |WP(e) - WP_\mathrm{baseline}(e)|
  \label{eq:equity_obj}
\end{equation}
linearized via a single auxiliary variable $v_\mathrm{eq}$ with
$v_\mathrm{eq} \geq WP(e) - WP_\mathrm{baseline}(e)$ and
$v_\mathrm{eq} \geq WP_\mathrm{baseline}(e) - WP(e)$ for all $e$.

\end{document}